# Automatic Detection of Solar Photovoltaic Arrays in High Resolution Aerial Imagery

Jordan M. Malof[1], Kyle Bradbury[2], Leslie M. Collins[1], Richard G. Newell[3]

[1]Department of Electrical & Computer Engineering, Duke University, Durham, NC 27708
[2]Energy Initiative, Duke University, Durham, NC 27708
[3]Nicholas School of the Environment, Duke University, Durham, NC 27708

*Abstract*— The quantity of small scale solar photovoltaic (PV) arrays in the United States has grown rapidly in recent years. As a result, there is substantial interest in high quality information about the quantity, power capacity, and energy generated by such arrays, including at a high spatial resolution (e.g., counties, cities, or even smaller regions). Unfortunately, existing methods for obtaining this information, such as surveys and utility interconnection filings, are limited in their completeness and spatial resolution. This work presents a computer algorithm that automatically detects PV panels using very high resolution color satellite imagery. The approach potentially offers a fast, scalable method for obtaining accurate information on PV array location and size, and at much higher spatial resolutions than are currently available. The method is validated using a very large (135 km²) collection of publicly available [1] aerial imagery, with over 2,700 human annotated PV array locations. The results demonstrate the algorithm is highly effective on a per-pixel basis. It is likewise effective at object-level PV array detection, but with significant potential for improvement in estimating the precise shape/size of the PV arrays. These results are the first of their kind for the detection of solar PV in aerial imagery, demonstrating the feasibility of the approach and establishing a baseline performance for future investigations.

*Index Terms*— solar energy, detection, object recognition, satellite imagery, photovoltaic, energy information.

## I. INTRODUCTION

The quantity of solar photovoltaic (PV) arrays has grown rapidly in the United States in recent years [2,3], with a large proportion of this growth due to small-scale, or distributed, PV arrays [4,5]. These small-scale installations are often found on the roofs of commercial structures, or private homes [4], and therefore are often referred to as rooftop PV.

Distributed PV offers many benefits [6], but integrating it into existing power grids is challenging. To understand and evaluate the factors driving distributed PV, and to aid in its integration, there is growing interest among government agencies, utilities, and third party decision makers in detailed information about distributed PV; including the locations, power capacity, and energy production of existing arrays. As a result, several organizations have begun collecting or publishing such information, including the Interstate Renewable Energy Council (IREC) [7], Greentech Media [8], and the US Energy Information Administration (EIA) [9][10].

Although the available information on distributed PV is expanding, it is nonetheless difficult to obtain. Existing methods of obtaining this information, such as surveys and utility interconnection filings, are costly and time consuming. They are also typically limited in spatial resolution to the state or national level [3],[6]. For example, the EIA began reporting state-level distributed PV data at the end of 2015 [9].

This work investigates a new approach for collecting distributed PV information that relies on using computer algorithms to automatically identify PV arrays in high resolution (≤ 0.3 meters per pixel) color aerial imagery. Fig. 1 is an example of 0.3 meter resolution imagery where the PV arrays have been annotated. At this resolution, it is possible to visually identify individual PV arrays, as well as their shape, size, and color. This permits the collection of distributed PV information at a very high geo-spatial resolution. Also, because the approach is automated, it is relatively inexpensive to apply (i.e., run a computer program), and to do so repeatedly as new imagery becomes available.

*A. Two challenges of collecting PV information in aerial imagery*

There are (at least) two major technical challenges to employing the proposed approach in a practical application. The first challenge involves developing a computer algorithm that can reliably identify the locations, shapes, or sizes of PV installations. The second challenge involves using the identified distributed PV imagery to infer the characteristics of the arrays, particularly power capacity and energy production. This information can then be aggregated into statistics for reporting.

In this work we address the first of these challenges, and present an algorithm for detecting PV arrays in aerial imagery, as well as estimating their shape and size. The main component of the algorithm is a supervised Random Forest classifier [11], which assigns a "confidence" to each pixel in an image indicating its likelihood of corresponding to a PV array. The performance of the algorithm is measured on a very large dataset of aerial imagery (135 km² area including more than 2,700 arrays) in which humans have annotated the PV array locations, shapes, and sizes. This dataset is part of a much larger dataset covering 900 km², including nearly 20,000 annotated PV arrays, and which is publicly available for download for comparing results [1]. Algorithm performance is evaluated based on correctly identifying PV pixels as well as identifying individual panel objects (and their precise shape and size).

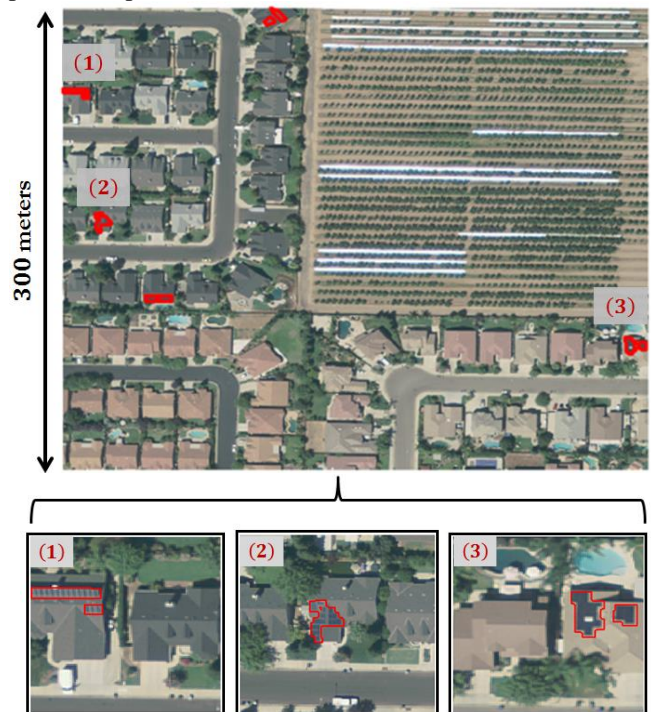

Fig. 1. An example of a color orthographic image (top) from the orthoimagery dataset, with human annotations shown in red outlines. Three of the annotations are enlarged in smaller images (bottom) so that the rooftop PV are more clearly visible.

### B. Related work: object detection in aerial imagery

The automatic detection of objects in aerial imagery (e.g., ortho-rectified imagery) has been researched extensively [12–15]. Some specific examples include the detection of roads [16–21], buildings [22–29], and vehicles [30–33]. In this published body of work, a large variety of algorithms have been proposed, employing techniques such as image processing, statistical modeling, machine learning classifiers, heuristic rules, and more.

The main component of the PV array detection algorithm proposed here is a supervised machine learning classifier called a Random Forest (RF) [11]. Supervised classifiers have previously been used for object recognition, including the RF [34,35], support vector machine (SVM) [31] and various types of neural networks [16,33,36]. The RF in particular has been applied for land cover classification [34] and object detection [35]. In [35] it was used to classify individual pixels into one of four classes: building, street, trees, and grass. The RF takes a similar role in this work, where it is used to classify individual pixels as PV, or not PV.

An important resource in aerial imagery recognition is a labeled dataset. Such datasets consist of imagery where each instance of the target object is indicated by a bounding box, polygon, or similar annotation. Such datasets are used for (i) the development of effective detection algorithms, and (ii) an accurate assessment of their performance. Ideally a labeled dataset will cover a large surface area, and have a large number of labeled objects. Large datasets better represent practical operating conditions, which involve large volumes of data that is collected for diverse environments and imaging conditions. Datasets used in recently published research typically include a few hundred labels, and a few hundred square meters of surface area [29,35,37,38]. Some specific recent examples include the SZTAKI-INRIA dataset for building detection, which includes 665 labeled buildings [37]. The OIRDS dataset for automotive detection consists of 1,800 labels [38]. In this work we utilize a dataset of color aerial imagery that encompasses 135 $km^2$ of area with more than 2,700 labels.

### C. Contributions of this work

The idea of automatically detecting PV arrays in aerial imagery was first investigated in a feasibility study [39] that employed a simple algorithm and a small dataset ($< 1 km^2$ area, with 53 PV array annotations). This work builds on that initial investigation with several contributions:

- A more sophisticated rooftop PV detection algorithm is developed, employing pixel-wise classification with an RF classifier, and post-processing steps that improve performance.
- The proposed algorithm is tested on a substantially larger dataset, covering 135 km$^2$, and including more than 2,700 PV array annotations. This dataset is also substantially larger than most datasets for similar object recognition tasks.
- The algorithm performance is measured at both a pixel level, and an object level. Unlike the previous study, the algorithm's ability to accurately measure both the shape and size of the target objects is assessed.
- The results are the first of their kind for PV array detection. Since the ground truth data are now publicly available [1], it is our hope that these findings serve as a baseline for further work.

The remainder of the paper is organized as follows. Section II describes the aerial imagery data that is used for algorithm development. Section III presents the proposed solar PV detection algorithm. Section IV presents the algorithm performance evaluation and Section V presents experimental results on the dataset. Section VI presents our conclusions and suggestions for future work.

## II. THE AERIAL IMAGERY DATASET

All experiments and algorithm development in this work utilize a large dataset of color (RGB) aerial imagery collected over the US city of Fresno, California. The imagery covers a total spatial area of 135 km$^2$. All of the imagery was collected in the same month in 2013, using aerial photography. The imagery has a spatial resolution of 0.3 meters per pixel, and all the imagery has been ortho-rectified. An example of the imagery is shown in Fig. 1, where the solar PV locations are annotated in red.

Further details about the data can be found at [1], where the data is also publicly available for download. The full imagery dataset is composed of 601 images that are each 5000 by 5000 pixels, across three cities, and with varying resolution. We chose to use imagery from Fresno, California because recently collected imagery was available (from 2013), with a high resolution ($0.3\ m$), and because Fresno has a large number of solar PV installations. Over 100 images of Fresno are available in [1], from which we randomly sampled 60 of the available images for the analysis presented here. The identification tags of these images are provided in the appendix for future investigations.

### A. Human annotations of true rooftop PV locations

In order to develop an effective computer vision algorithm, as well as accurately assess its performance, it is necessary to have the precise locations where PV installations appear in the aerial imagery. In order to obtain this information, human observers visually scanned the imagery and annotated all of the (visible) PV arrays. For improved quality, two annotators scanned each part of the imagery, and their annotations were combined by taking a union of each observer's annotations. There were a total of 2,794 individual solar PV regions in the imagery after the merging process. Note again that this is a subset of the 19,863 annotations available in [1]. Some examples of annotated regions are shown at the bottom of Fig. 1 and in Fig. 4a.

To avoid a positive bias in the performance evaluation of the proposed detection algorithm, we split the available imagery into two disjoint datasets: Fresno Training and Fresno Testing. This is a common approach for validating supervised machine learning algorithms, such as the RF model used in our detection algorithm. A summary of the imagery in each dataset is presented below in Table 1. The data was split between training and testing at a ratio of 2:1, in order to provide enough solar array examples to effectively train the RF model (see Section III.C for details about the RF).

TABLE 1
SUMMARY OF FRESNO COLOR ORTHOIMAGERY DATASET

| Designation | Area of Imagery | Number of PV Annotations |
|---|---|---|
| Fresno Training | $90\ km^2$ | 1780 |
| Fresno Testing | $45\ km^2$ | 1014 |

## III. THE PROPOSED PV DETECTION ALGORITHM

In this section we present the details of the proposed solar PV detection algorithm. We begin with a brief overview of the primary processing steps, followed by individual sections providing more details about each step.

### A. Algorithm overview

The proposed rooftop PV algorithm takes RGB color aerial imagery as input and performs four major processing steps, as illustrated in Fig. 2.

1) Feature extraction. This step consists of extracting image statistics, or features, around each pixel that characterize the colors, textures, and other patterns surrounding the pixel. The feature extraction step effectively maps the 3-channel RGB image into an M-channel image, where $M$ is the number of features extracted around each pixel location.

2) Random Forest Classifier. The image statistics computed in the feature extraction stage are the input to a trained RF classifier. The RF is a machine learning classification model that assigns a probability, or "confidence", to each pixel in the imagery. The confidence value indicates how likely the pixel is to correspond to a PV array. The output of this step is a single channel image, or spatial map, of where PV arrays are likely to be located. An example image and associated confidence map are shown in Fig. 4.

3) Post-processing. This step is designed to improve the accuracy of the confidence map that was generated in the RF classification step. This process consists of identifying high confidence individual pixels (i.e., local maxima locations) and then growing regions of pixels around them. All pixel confidence values outside of these grown regions are then set to zero.

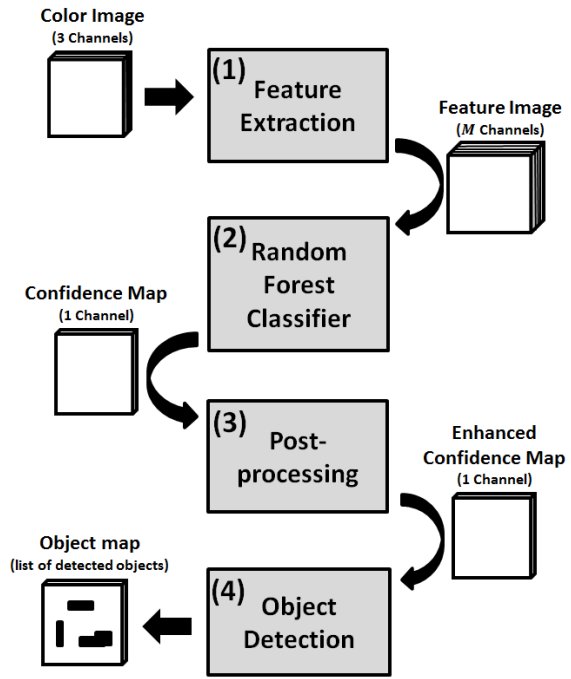

Fig. 2. A flowchart of the PV detection algorithm. Each gray block corresponds to a major processing step. Additionally, the input and output of each stage is also shown on the right or left of each block.

4) Object detection. This step identifies groups of contiguous high confidence pixels that are likely to correspond to a single PV array. Each identified group of contiguous pixels is returned from this step as a detected object, and the confidence of that object is set to the value of the maximum pixel confidence value in that object. The output of this step is a list of objects and their confidence values, which is used to perform object-based scoring.

### B. Feature extraction

The feature extraction process takes a 3-channel RGB aerial image as input, and returns an M-channel "feature image". The feature image, as it is called here, can be thought of as a new representation of the original image, where each pixel in the original image is now represented by an $M$-dimensional vector of feature values that are computed using the original RGB image. An important consideration in this work is computational efficiency, so that the proposed algorithm can be applied at a national scale. It should therefore be possible to compute the features quickly, with few computations. One class of image features that achieves this objective consists of image statistics (e.g., means and variance) computed in rectangular windows. These features can be computed very efficiently using integral images [40,41].

In the proposed feature extraction approach, each pixel is represented by the means, $\mu$, and variances, $\sigma^2$, computed in several 3x3 windows surrounding it, for each channel of the RGB image. A window size of 3x3 was chosen because it roughly corresponds to the size of the smallest PV array in the dataset. A larger choice of window size would risk mixing PV pixels with background pixels, and thereby obscuring the information useful for identifying individual PV panels. A smaller window size would be too small to compute any statistics. Note that each window results in 6 features total (2 features for each of the 3 RGB image channels).

The feature set consists of features extracted from several of these windows surrounding a center pixel, $p_0$, which is illustrated in Fig. 3. The windows are organized into two rings around $p_0$, and each ring consists of 9 windows. The goal of this structure is to capture image statistics in the area surrounding the pixel location. To precisely describe the extraction location of each window, we can characterize each window's location by its vertical offset, $y$, and horizontal offset, $x$, from $p_0$. This is illustrated in Fig. 3. A set of 9 windows (one ring) is denoted here by $S_r$, where the subscript $r$ parameterizes the locations of the windows in the ring. The locations of the windows in $S_r$ are then given by

$$S_r = \{(x,y): x \in \{0,-r,r\}, y \in \{0,-r,r\}\}. \quad (1)$$

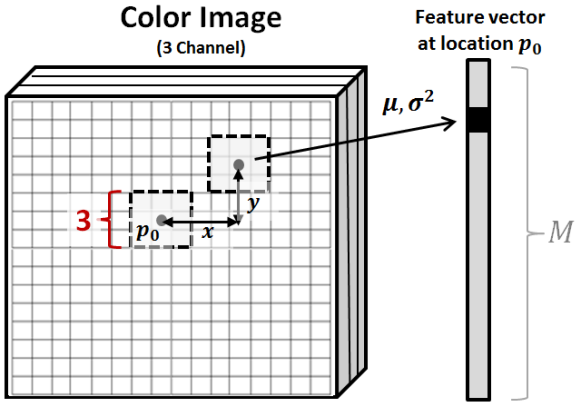

Fig. 3. Illustration of pixel-based feature extraction at a single pixel location, $p_0$. Features are extracted in several windows around $p_0$, and each window is 3x3 pixels in size. Each window is described by its horizontal and vertical offset from $p_0$, given by $x$ and $y$ respectively. A mean, $\mu$, and variance, $\sigma$, are computed in each window, and across each of the three channels of the image (so 6 total features for each window). All the features are combined into a single $M$-dimensional vector representing the pixel at $p_0$.

There are 9 windows in each ring, and each window yields 6 features, resulting in 54 total features per ring. In this work, we extracted features in two rings, given by $S_2$ and $S_4$. Note that $S_2$ and $S_4$ share a window location at $(x,y) = (0,0)$. One of these duplicates is removed, leaving a total of 54+54-6=102 total features. These two rings, $S_2$ and $S_4$, were found to provide a good compromise between measuring useful local image statistics, and increasing the computation time due to increases in the feature-space size.

### C. The Random Forest Classifier

Random Forests (RFs) [11] are a state-of-the-art supervised statistical classification method. They have been successfully applied to a variety of problems, such as image processing [42], medical diagnosis [43], pose recognition [44,45], and remote sensing [34,35]. In this work we use the RF to classify each pixel in the imagery, similar to the way it was used in some other contexts [35,45]. The RF receives the feature vector for each pixel and assigns it a probability, or confidence, indicating the likelihood that it corresponds to a PV array.

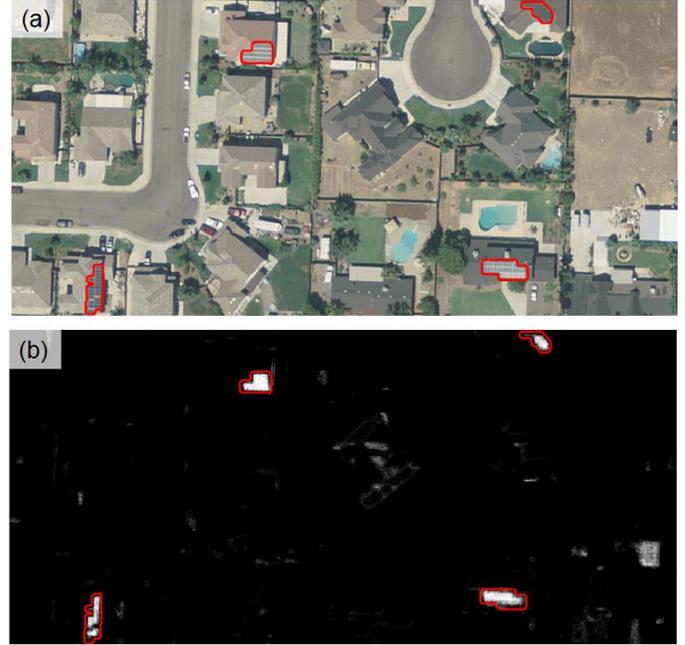

Fig. 4. An example of an aerial image (top) and its corresponding confidence map (bottom). In both images the true solar PV locations have been annotated in red. The confidence map is the output of stage two in the detection algorithm (see Fig. 2).

Although the RF has many advantages, there are two primary reasons the RF was employed in this work. First is the ability of the RF to learn complex nonlinear relationships between input and output variables. This is important because the relationship between the image features and the pixel labels (i.e., PV or non-PV) are very complex. The second motivation for the RF is its known computational speed during training and testing [34,35,45]. It can also be implemented on graphics processing units for further speed improvements [41]. This computational efficiency is important for handling the massive amounts of data common in high resolution aerial imagery applications. For example, our datasets consists of 1.5 billion pixels, however this encompasses only 135 km$^2$ of the United State's nearly 9.857×10$^6$ km$^2$ area.

The RF actually consists of an ensemble of $T$ simpler supervised classifiers called decision trees [46]. An illustration of an RF is provided in Fig. 5. Each tree consists of a series of decision nodes which terminate (at the bottom of the tree) in a leaf node. To classify a new feature vector, $x$, it is presented to the top decision node, and it is subsequently directed down the tree, based on the values in the feature vector, until it reaches a leaf node. At the leaf node a probability is assigned to the vector indicating to which class (e.g., PV or non-PV) it belongs.

During training, each tree is "grown" independently of the other trees, in a top-down manner, using a random bootstrap sample of pixels from the training data. The decision nodes are learned such that they best separate the training data according to some performance measurement (e.g., the Gini impurity index or information gain). In this work we use the Gini index. Each node of each tree considers only a random subset of the input features (of size $m$) when inferring how to split the data. The parameter $m$ is often cited as the only major adjustable parameter of the RF, and a conventional setting that usually works well is $m = \sqrt{M}$, where $M$ is the number of feature dimensions [34].

Decision nodes are created until (i) splitting no longer improves the Gini index or (ii) it would result in fewer than 5 observations in a leaf node. The parameter settings for the RF, and other algorithms in this work, are presented in Table 3.

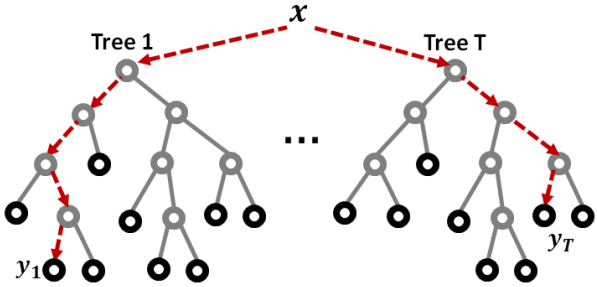

Fig. 5. Illustration of a RF classifier architecture. The RF consists of several decision trees, where each tree consists of decision nodes (blue circles) and leaf nodes (green nodes). To classify an input feature vector, $x$, it is presented to each tree independently. For a given tree, $x$, is passed down the decision nodes based on its values until it reaches a leaf node. At the leaf node it receives a probability indicating to which class it belongs. Here $y_t$ indicates the probability of belonging to the PV class. The RF output is creating by averaging the probabilities returned from each tree.

*D. Post-processing*

The goal of the post-processing (PP) step is to improve the pixel-wise classification accuracy of the raw confidence maps, as well as to make them better suited for the object detection step. The algorithm for PP is outlined in Table 2, and an example of the input and output of this process is shown in Fig. 6.

Broadly speaking, the PP algorithm identifies individual high confidence pixels (local maxima) and then grows a new smooth region (in terms of confidence values) around them. Local maxima are retained for region growing only if they (i) exceed some minimum threshold, $c_0$, and (ii) they are the largest maxima in a surrounding square window of length $L_s$. This filtering criterion is designed to remove maxima locations that are likely to be false alarms. Regions are grown around each remaining maxima location using Otsu's method [47], which automatically segments pixels into foreground (high confidence) and background (low confidence) regions. The region growing takes place in a square window of length $L_g$ that is centered on each maxima location.

TABLE 2

**The Post-processing Algorithm**

**Data**: $I \equiv$ Confidence Map, **Parameters**: $R_c, L_g, L_s, c_0$

**Result**: $I' \equiv$ Enhanced Confidence Map

**Initialize:** $I' \leftarrow$ image of zeros, with size equal to $I$

1. Apply non-maximum suppression to $I$, using filters that are $L_s$ x $L_s$ in size
2. Remove all maxima below global threshold, $c_0$
3. **For** each remaining maxima location, $i$, **do**
    (a) Crop an $L_g$ x $L_g$ region around $i$
    (b) Use Otsus method [47] to find foreground pixels in the cropped region
    (c) Find all connected components
    (d) Retain only the connected component, C, that contains the maxima location, $i$
    (e) Place C into image $I'$, with all of the pixel values in C set to $I(i)$
   **end**
4. Apply a morphological closing and dilation [48] to $I'$, with a disk of radius $r_1$ and $r_2$, respectively

The output of this PP operation, $I'$, consists of many small connected regions that all have the same confidence value. This makes object extraction easier. The final step of post-processing is to apply morphological closing and dilation [48], respectively, in order to smooth the grown regions. These operations are performed with a disk with radius $r_1$ and $r_2$, respectively. The parameter values of the PP algorithm are provided in Table 3. These values were set by optimizing the performance of the algorithm on the Fresno Training data (see Section IV.C).

*E. Object Detection*

The object detection phase identifies groups of adjacent, or neighboring, high confidence pixels and identifies them as detected objects. This is achieved by first thresholding the confidence map: any pixel confidence greater than zero is set to one, and all others are set to 0. The result is a binary image, which is used for finding contiguous groups of pixels. The resulting connected regions are all taken as detected objects. The confidence of each region is given by the maximum confidence pixel in that region. An example output from this process is shown in Fig. 6.

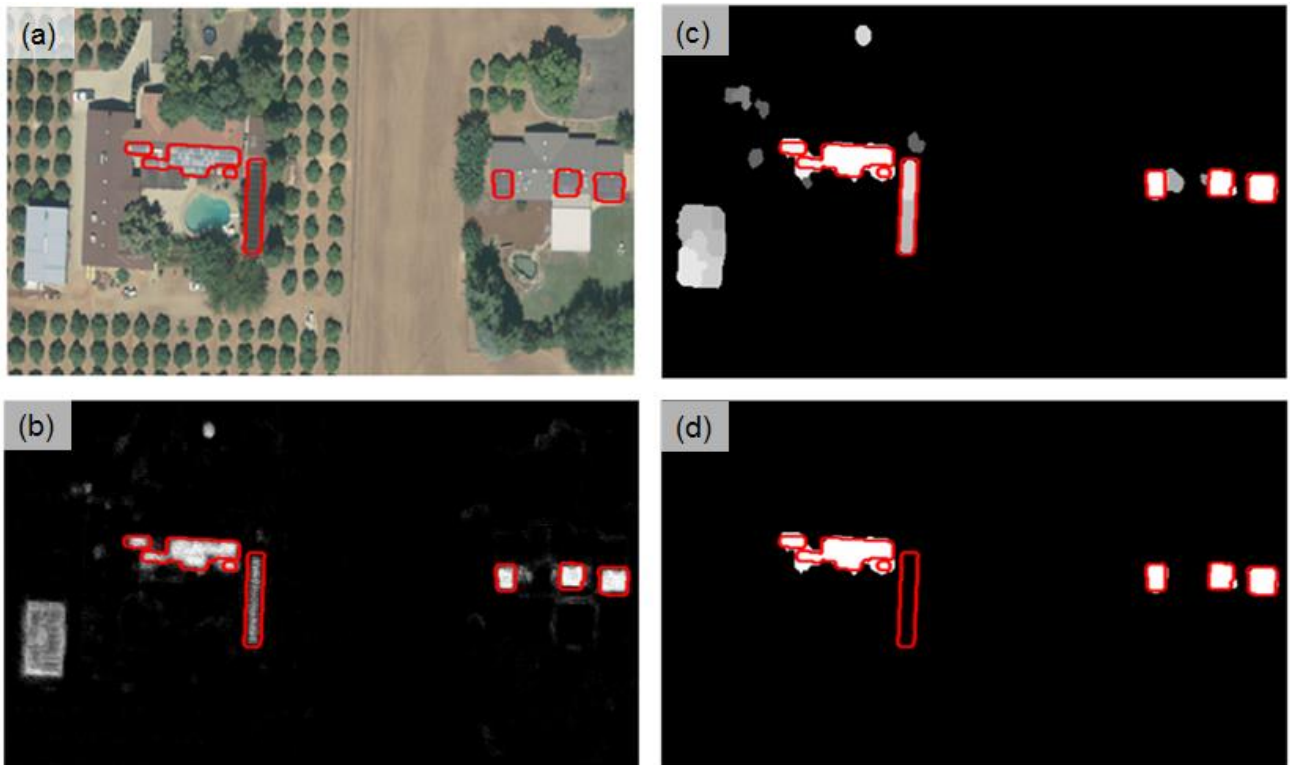

Fig. 6. Example output of the rooftop PV detection algorithm after several of the major processing steps. Four different images are shown, (a)-(d), and each image shows the human PV annotations in red. (a) is the original RGB image. (b) is the confidence map output from the Random Forest classifier, without post-processing; brighter pixels indicate higher confidence. (c) is the confidence map after post-processing. (d) shows the objects detected after the object detection stage of processing. Given the detection rate and false alarm rate employed in this example, the detector correctly removes all of the false alarms, while losing one of the true panel regions.

## IV. EXPERIMENTAL DESIGN

This section begins with an overview of the experimental design used for the experiments in Section V, followed by a description of our performance evaluation methodology.

We conducted two experiments, with the primary goal of measuring the performance of the proposed PV array detection algorithm. The first experiment measures how well the algorithm identifies individual PV pixels: pixel-based classification performance. The second experiment measures how well the algorithm can identify objects (groups of pixels) that correspond to PV array annotations, as well as their precise shape and size. The experiments are conducted on two datasets of aerial imagery denoted as Fresno Training, and Fresno Testing (see Table 1). As described in Section II, all PV arrays visible in the imagery were annotated by humans to provide ground truth pixels/objects for use in scoring the detector.

The primary role of the Fresno training dataset was to train the RF classifier, as well as optimize other parameters associated with the detection algorithm. The Fresno Testing dataset was used to obtain an unbiased performance estimate for the detector. This is a common approach for supervised machine learning algorithms [49]. The performance metric used to evaluate the performance of the algorithm is the precision recall (PR) curve. The PR curve is a popular performance metric for object detection in aerial imagery [16,22,50,51], and therefore it is adopted here.

The next few sections describe PR curves, the Jaccard index (used to measure the degree of overlap between detected objects with human annotations), and how the algorithm parameters were optimized.

### A. Performance metrics

PR curves measure the performance tradeoff between making correct detections and false detections, as the sensitivity of a detector, or classifier, is varied. An illustration of a PR curve is shown in Fig. 7. The x-axis of a PR curve is the recall, $R$, which is the proportion of all true target objects (e.g., PV arrays) in the data that were returned by the algorithm as detections. The y-axis is the precision, $P$, which is the proportion of all detected objects (i.e., both true and false) which are true targets. An effective detector will tend towards the top right corner of the PR curve, thereby maximizing both recall and precision. A detector that detects objects randomly (i.e., it is ineffective) will achieve a precision that is equal to the proportion of objects in the dataset that are targets. For example, in the pixel-based PV detection experiments, roughly 0.07% of the Fresno Testing pixels correspond to PV arrays. Therefore a random detector would achieve $P = 0.0007$, for all values of $R$.

The sensitivity for a given detection algorithm can be varied by raising or lowering a threshold, $t_0$, that is applied to the confidence values of the list of potential detections (e.g., pixels or objects). All potential detections above $t_0$ are accepted as detections, and all potential detections below $t_0$ are rejected. P and R are then computed based on the group of accepted detections.

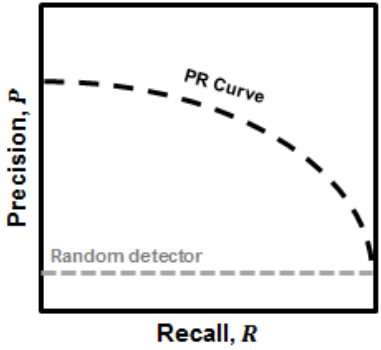

Fig. 7. Illustration of a PR curve. A good detector will obtain a curve that is closer to the top right corner of the graph. Random guessing results in a line that achieves constant precision for all values of R, where the precision is equal to the proportion of total detections returned by the algorithm that are from the target class (e.g., PV arrays).

### B. Linking detections to human annotations

One issue that arises with object-based scoring is determining when a detected object should be considered a correct detection. A detected object (i.e., a region labeled as a PV array) may overlap with a PV annotation from the ground truth data, but how much overlap should be required to declare it as detected correctly? This problem is apparent in Fig. 6 Fig. 6d, where none of the detected objects match *perfectly* with the human annotations, but they might reasonably be considered correct detections. To address this issue, a metric is needed to measure the shape/size similarity between two objects (i.e., groups of pixels).

One metric that has been utilized for this purpose is the Jaccard index [52]. The Jaccard index, $J$, for two image objects (groups of pixels), denoted $A$ and $B$ respectively, is given by

$$J(A,B) = \frac{|A \cap B|}{|A \cup B|}. \quad (2)$$

Fig. 8 below shows the Jaccard index for two objects as their level of overlap varies. The Jaccard index allows us to measure precisely how similar a detected object is to a human annotation. A threshold can then be applied where only detected objects above the threshold (with respect to a human annotation) can be considered a correct detection. Ultimately, the choice for this threshold should depend on the final application of the detector, and the corresponding level of shape/size accuracy that is needed. Therefore, we report object-based performance for multiple thresholds. A similar approach was recently adopted in [53] for building detection.

In some instances a detected object will overlap with multiple ground truth annotations (see the left-most three annotations in Fig. 6d for an example). In this case the multiple annotations are treated as one annotation composed of the union of the individual annotations. If the union of the annotations has a sufficiently high Jaccard index with respect to a detection, all three annotations are considered to be detected by the detector.

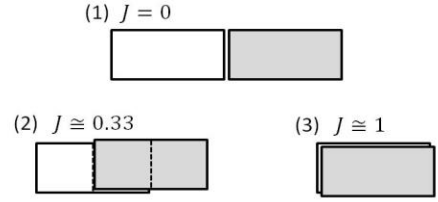

Fig. 8. Illustration of the Jaccard Index, $J$. The gray and white boxes represent two sets of pixels, such as a detected object from the solar PV detector, and a true solar PV annotation. As the degree of overlap of the two sets increases, $J$ increases from 0 toward 1.

### C. Algorithm training and optimization

All training and parameter optimization was performed on the Fresno Training dataset. The final set of chosen parameters for the algorithm is shown in Table 3. These parameters are used in all experiments.

The RF classifier itself was trained using five million pixels from the Fresno Training dataset. This subset of pixels was chosen by first selecting all of the available solar PV pixels (roughly 500,000), and then randomly sampling the remaining non-PV pixels from the training imagery. Using increasing numbers of pixels improves performance but at the cost of increasing the computation time of the RF. Five million was found to be a good tradeoff between performance and computation time on the training data.

The parameters were chosen in order to optimize performance on the training data. This parameter optimization was done by measuring the performance of the algorithm (on the Fresno Training dataset) as the parameters were varied over a coarse grid of potential values. Note that the parameter $m$ was set to the conventional value of $\sqrt{M}$, rather than being optimized.

TABLE 3
DETECTION ALGORITHM PARAMETERS

| Symbol | Processing Step | Quantity | Value |
|---|---|---|---|
| $T$ | RF | Number of RF trees | 30 |
| $M$ | RF | Number of features in RF | 102 |
| $m$ | RF | Number of variables to sample at each tree node in RF | $\sqrt{M} \cong 10$ |
| $L_s$ | PP | Length of the of square non-maximum suppression region | 9 |
| $c_0$ | PP | Global confidence threshold after non-max suppression | 0.375 |
| $L_g$ | PP | Length of the square region around each maxima location used for Otsu's method | 19 |
| $r_1, r_2$ | PP | Disk radius used for morphological closing, and dilation, respectively | 5,2 |

## V. EXPERIMENTAL RESULTS

This section describes the experimental results obtained using the experimental design discussed in Section IV. First,

pixel-based performance results are presented, followed by object-based performance.

*A. Pixel-based performance*

The pixel-based performance for the PV detection algorithm, on both the training and testing data, is shown in Fig. 9. Results are shown for the RF, and the RF after PP has been applied (RFPP). The primary goal of this experiment was to demonstrate that the RFPP algorithm can effectively detect PV array pixels. The results on the Fresno Testing dataset provide an unbiased estimate of the performance of the RF and RFPP algorithms. The results indicate that the solar PV detector is very effective at discriminating non-panel pixels from panel pixels. This is made most clear by considering how well a random detector (i.e., a completely ineffective detector) would perform. Recall from Section IV.A that, because PV arrays constitute only 0.07% of the pixels in Fresno Testing, the random detector achieves $P = 0.0007$ for all values of $R$. Both the RF and RFPP detectors achieve performance far above this.

Further insight can be obtained from the results in Fig. 9 by comparing the performance of the detectors on the training data and testing data, respectively. As is expected, the results indicate that there is an overall performance drop between the training data and the testing data. Quantitatively this means that, for each value of $R$, the algorithm typically obtains a lower $P$ on the testing data than it does on the training data. One exception to this occurs for the RFPP algorithm when $R$ is below 0.6, however the testing and training performance is similar at these operating sensitivities.

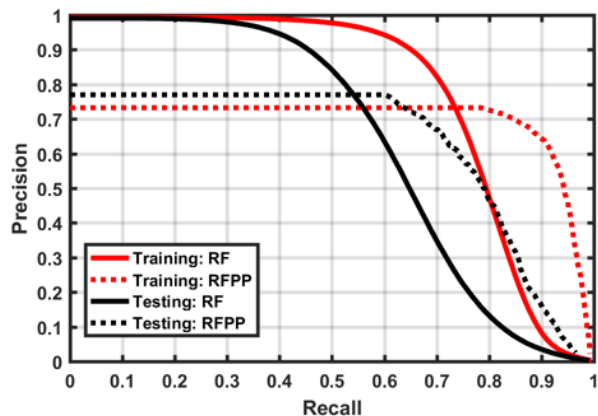

**Fig. 9. PR curves for the pixel-wise performance of the PV detector on the Fresno Training dataset (red), and the Fresno Testing dataset (black). For each dataset, the performance of the detector is shown before post-processing (solid lines), and after post-processing (dashed lines. The random detector for this problem achieves $P = 0.0007$ for all values of $R$, but this is not shown due to its small magnitude.**

The results also suggest that the main contributor to the performance loss incurred on the testing data is the RF classifier (as opposed to RFPP). This is because the RF algorithm performance drops between the training and testing dataset, however, the RFPP algorithm offers the same advantages on both the training and testing dataset; relative to the performance of the RF alone. This suggests that the RF is overfitting to the training data, or in other words, the RF learned to recognize patterns that are too unique to the training data, and as a consequence it less effectively identifies previously unseen PV arrays in the testing data. This can be addressed in many ways, and is an important consideration for future work.

*B. Object-based performance*

The primary goal of this experiment is to demonstrate the effectiveness of the detector. Further, we want to examine how well the detector can identify the precise shape/size of individual PV arrays. As a result, we measure the object-based performance of the detector on the Fresno Testing dataset for varying settings of the Jaccard index during scoring. These resulting PR curves are shown in Fig. 10.

The results indicate that the object-based performance of the detector is once again well above that of the baseline random detector performance. Although this is true for all values of $J$, the performance of the detector decreases rapidly as $J$ increases. As a specific example, when $J = 0.1$ the detector achieves $R = 0.7$ with $P = 0.6$, while at $J = 0.5$, $R = 0.55$ at the same value of $P$. When $J = 0.7$, the detector never reaches $P = 0.6$. This outcome is expected because, as $J$ is increased, many of the objects detected that are near true PV locations are no longer considered correct detections. This also results in more PV annotations remaining undetected, even when the detector is operated with high sensitivity. This is why the maximum $R$ obtained for each detector decreases as $J$ increases.

Different values for $J$ are likely to be appropriate depending on the intended purpose of the detector. For example, lower $J$ values (e.g., $J = 0.1$) are appropriate for applications where only the general location of target objects is important, and obtaining the precise shape/size is not. In the context of solar PV array detection, this may be the case if the detector is used as a preprocessing step for further, and more sophisticated (but slower), detection algorithms. Note that when operated with $J = 0.1$ the detector is capable of detecting roughly 90% of the targets, with $P \cong 0.1$. Since there are roughly 1000 PV arrays in the testing data, this corresponds to roughly 10000 total detections returned by the detector (900 true detections and 9,100 false detections) over the $45 \text{ km}^2$ testing area. This dramatically reduces the amount of image locations that must be considered for further processing, facilitating the use of more sophisticated subsequent processing. The detector proposed here is designed to operate quickly on large datasets, and therefore could be used in this role.

In contrast to lower $J$ values, a higher value (e.g. $J = 0.7$) is appropriate for detection applications where it is important to accurately estimate the size and shape of target objects. In the context of solar PV array detection, this may be the case, for example, if the goal is to estimate the power capacity of individual solar PV arrays. Setting $J$ to higher values will lead to a performance measure that better reflects the capability of a given detector to achieve that goal, which is a much more difficult task than simply detecting the likely presence of an object (using, e.g., $J = 0.1$). This difficulty is reflected in the much poorer performance of the proposed detector on this task (e.g., see Fig. 10 with $J = 0.7$). Looking forward, the performance reported here for $J = 0.7$ establishes a baseline for future improvement in achieving this type of goal.

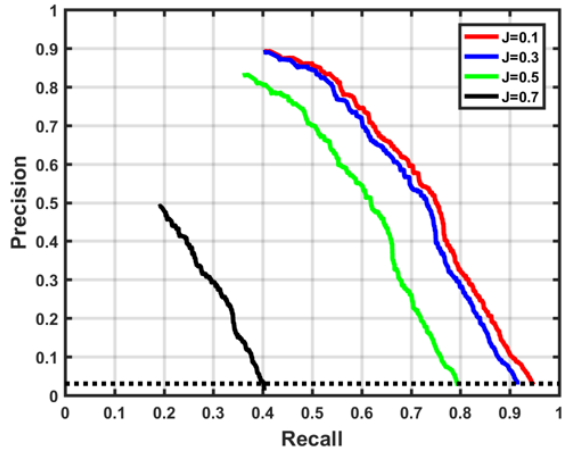

Fig. 10. PR curves for the object-based performance of the rooftop PV detector on the Fresno Testing dataset. Each PR curve corresponds to a different setting of the Jaccard index, *J*, during scoring. The left-most point of the curves represents the performance when classifying every object with confidence of one (i.e., the maximum RF output) to a detection. With object-based scoring, the detectors are not guaranteed to place objects over all true PV array locations, and indeed, none of the detectors reach $R = 1$.

## VI. CONCLUSIONS AND FUTURE WORK

We investigated a new approach for the problem of collecting information for small-scale solar PV arrays over large areas. The proposed approach employs a computer algorithm that automatically detects solar PV arrays in high resolution ($\leq 0.3\ m$) color (RGB) imagery data. A detection algorithm was developed and validated on a very large collection of aerial imagery ($\geq 135 km^2$) collected over the city of Fresno, CA. Human annotators manually scanned and annotated solar PV locations to provide ground truth for evaluating the performance of the proposed algorithm. Performance was measured in a pixel-based and object-based manner, respectively, using PR curves. In the case of object-based scoring, the algorithm was also scored based on how well it can identify the shape and size of the true panel object.

The results demonstrate that the algorithm is highly effective on a per-pixel basis. The PR measures indicate it can detect most of the true PV pixels while removing the vast majority of the non-PV pixels. The object-based PR curves indicated that the algorithm was likewise effective at object detection, however, it was far less effective at estimating the precise shape/size of the PV arrays.

The results presented here are the first of their kind for distributed PV detection in aerial imagery, and demonstrate the feasibility of collecting distributed PV information over large areas using aerial or satellite imagery. This may ultimately yield a faster, cheaper, and more scalable solution for the large scale collection of distributed PV information, and potentially information for other aspects of energy production and consumption as well. While the results here demonstrate the promise of this approach to information collection, several challenges remain as opportunities for future work.

### A. Future work

*1) Improved detection algorithms*

Because the results here are the first of their kind for this problem they establish a baseline performance, or benchmark, for future algorithm development. To facilitate such efforts, the data used in this work is freely available for download [1], and the exact images used in our experiments are listed in the supplemental materials. It is our hope that others will build upon these results, and develop increasingly effective detection algorithms.

*2) Inferring capacity and energy production*

Another important line of future work is the inference of PV array capacity, energy production, and other characteristics from the imagery. Recall from the Section I, that this is the second major technical challenge for creating a complete system for extracting PV information from aerial imagery. This second challenge could be pursued using the imagery detected from the PV detector, or otherwise using the ground truth annotations in the aerial imagery dataset.

*3) Establishing practical performance needs*

While the results here demonstrate the ability of an algorithm to discriminate between PV and non-PV imagery (as compared to a random detector), it is unclear what levels of performance would be needed for different practical applications. Creating a complete system for inferring distributed PV information would help reveal what level of performance is needed from the detection stage in order to obtain practically useful energy information; with qualities and reliability that is similar, or better than, current estimation strategies (e.g., the EIA [9]).

*4) Information collection for other energy resources*

Finally, we hope this work also motivates the collection of other energy information from aerial imagery, in addition to distributed PV. Other examples might include inferring the energy consumption of individual households, and (from that information) counties, or cities. This could be pursued, for example, by estimating the volume of a household from aerial imagery. Other elements of the energy system could also potentially be detected, such as power lines, or power plants.

## ACKNOWLEDGEMENTS

This work was supported in part by the Alfred P. Sloan Foundation and the Wells Fargo Foundation. The content is solely the responsibility of the authors and does not necessarily represent the official views of the Alfred P. Sloan Foundation or the Wells Fargo Foundation

**SUPPLEMENTAL MATERIALS**

Training Image Tags:
11ska505665
11ska580710
11ska475635
11ska580860
11ska475875
11ska565845
11ska565905
11ska490860
11ska325740
11ska460725
11ska490605
11ska430815
11ska400740
11ska580875
11ska655725
11ska595860
11ska460890
11ska655695
11ska640605
11ska580605
11ska595665
11ska505755
11ska475650
11ska595755
11ska625755
11ska490740
11ska565755
11ska520725
11ska595785
11ska580755
11ska445785
11ska595800
11ska625710
11ska520830
11ska640800
11ska535785
11ska430905
11ska460755
11ska505695
11ska565770

Testing Image tags:
11ska625680
11ska610860
11ska445890
11ska520695
11ska355800
11ska370755
11ska385710
11ska550770
11ska505740
11ska385800
11ska655770
11ska385770
11ska610740
11ska550830
11ska625830
11ska535740
11ska520815
11ska595650
11ska475665
11ska520845